\documentclass{article}
\usepackage{amssymb}
\usepackage{multirow}
\usepackage{booktabs}
\usepackage{graphicx}
\usepackage{wrapfig} 
\usepackage[final]{corl_2025} 

\title{SimShear: Sim-to-Real Shear-based Tactile Servoing}

\author{
\textbf{Kipp McAdam Freud}\textsuperscript{*}, \hspace{1mm}
\textbf{Yijiong Lin}\textsuperscript{*}, \hspace{1mm}
\textbf{Nathan F. Lepora} \vspace{1mm}\\
School of Engineering Mathematics and Technology, University of Bristol \\ Bristol Robotics Laboratory, University of Bristol \vspace{0.5mm}\\ 
\vspace{-7mm}
}

\begin{document}
\maketitle

\renewcommand{\thefootnote}{\fnsymbol{footnote}}
\footnotetext{\textsuperscript{*} These authors contributed equally. Correspondence to  \href{mailto:n.lepora@bristol.ac.uk}{\fontfamily{qcr}\selectfont n.lepora@bristol.ac.uk}}

\begin{abstract}
We present SimShear, a sim-to-real pipeline for tactile control that enables the use of shear information without explicitly modeling shear dynamics in simulation. Shear, arising from lateral movements across contact surfaces, is critical for tasks involving dynamic object interactions but remains challenging to simulate. To address this, we introduce shPix2pix, a shear-conditioned U-Net GAN that transforms simulated tactile images absent of shear, together with a vector encoding shear information, into realistic equivalents with shear deformations. This method outperforms baseline pix2pix approaches in simulating tactile images and in pose/shear prediction. We apply SimShear to two control tasks using a pair of low-cost desktop robotic arms equipped with a vision-based tactile sensor: (i) a tactile tracking task, where a follower arm tracks a surface moved by a leader arm, and (ii) a collaborative co-lifting task, where both arms jointly hold an object while the leader follows a prescribed trajectory. Our method maintains contact errors within 1–2 mm across varied trajectories where shear sensing is essential, validating the feasibility of sim-to-real shear modeling with rigid-body simulators and opening new directions for simulation in tactile robotics.\\ Project webpage: \url{https://yijionglin.github.io/simshear/}
\end{abstract}



\section{Introduction}

The application of learning algorithms within robotics has the potential to facilitate complex and generalized behaviours that are challenging to achieve using traditional control methods. However, such learning algorithms often require large datasets, necessitating resource-intensive data collection efforts~\cite{ibarz2021train}. To address this, recent studies have used data obtained through physics-based simulations, with robot learning to transfer control policies to real environments~\cite{peng2018sim, zhao2020sim, liu2020real}. 

However, physics engines simplify real-world complexities to reduce computational demands, leading to a ``sim-to-real gap'' when transferring control policies. Specifically, in tactile robotics, simulations of the interaction dynamics between objects and soft tactile sensors typically employ rigid bodies with penetrable contacts to approximate the deformation experienced by actual tactile sensors. This excludes the use of shear in the trained policies, thereby restricting the scope of tasks achievable through sim-to-real training to those using contact depth and geometric information~\cite{salvato2021crossing, stoffregen2020reducing}.
\begin{figure*}[]
\centering
\includegraphics[ trim=35 500 40 0, clip, width=\textwidth]{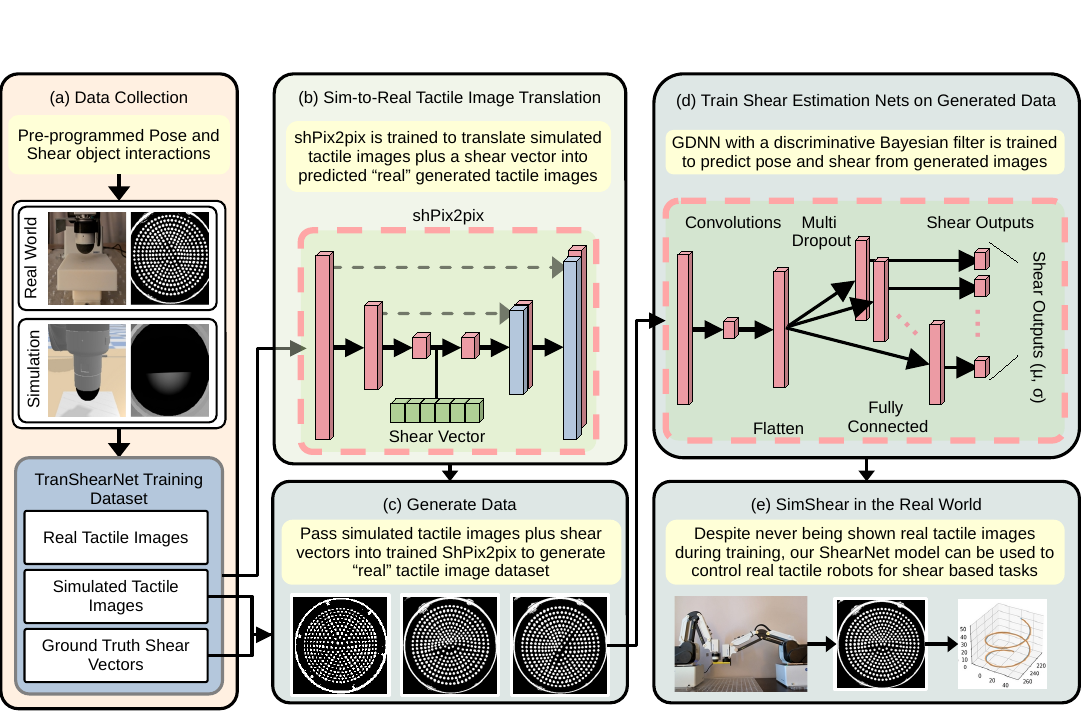}
\caption{Overview of SimShear: our shear-based Sim-to-Real pipeline for tactile robotics.}
\label{fig:hero}
\vspace{-1.5em}
\end{figure*}
Current approaches to transferring tactile sim-to-real policies typically involve a real-to-sim pipeline \cite{yang2023sim, lin2022tactile}, where a policy is trained with simulated data, then incoming real-world data is converted real-time into a simulated equivalent to use in the trained policy for action generation. This computationally-expensive design choice is due to a one-to-many sim-to-real gap where the simulated tactile images correspond to many real tactile sensor images containing identical geometric and contact depth information but arbitrary amounts of shear; as the inverse sim-to-real transformation is undefined, by necessity a real-to-sim pipeline is used even though it is expensive at run-time. 

This research aims to improve tactile sim-to-real policy transfer by leveraging contact data available in most rigid-body physics simulators to build sim-to-real image translators that include shear information and return estimated `real' tactile images that correctly contain shear deformations. By transforming simulated images into their real analogs during training, it becomes possible to build policies that operate purely on real tactile images, thereby removing the need for a real-to-sim transformation at each deployment step. This streamlined process reduces computational overhead and simplifies the overall control pipeline. Our main contributions are:

\noindent 1. We introduce shPix2pix: a conditional U-Net GAN architecture that incorporates shear information into simulated tactile sensor images for image-to-image translation. By modeling deformations due to lateral displacements, shPix2pix enables the generation of realistic tactile images that contain shear deformations not modeled by our rigid-body simulator. 

\noindent 2. We train a ShearNet: a Gaussian Density Neural Network (GDNN) that leverages the shPix2pix-generated shear-based tactile images to estimate both contact pose and shear. We validate GDNN performance when trained on `real' shPix2pix generated tactile images and show this significantly outcompetes a baseline using standard pix2pix-generated tactile images.

\noindent 3. We demonstrate SimShear with two control tasks involving a pair of low-cost desktop robotic arms: a collaborative tactile tracking task and a collaborative co-lift task. Our results demonstrate how our shear-based sim-to-real approach enables manipulation performance using shear and validates that shear-aware models trained in simulation can effectively transfer to real environments.

We will open-source all code and the designs of the tactile sensors necessary to reproduce this work.

\section{Related Work}

Many studies have reported promising results involving tactile sensing in simulation~\cite{kim2023marker,luu2023simulation,xu2023efficient,si2024difftactile,zhao2024fots,du2024tacipc}, using a range of rigid-body and soft-body simulation methods. For example, Ding et al. used a zero-shot sim-to-real approach with domain randomization to complete a door-opening task~\cite{ding2021sim}. Wu et al. (2019) utilized the tactile sensor on a Barrett Hand to optimize grasping during object manipulation \cite{huang2019learning}. Si et al. achieved zero-shot sim-to-real grasp stability prediction task~\cite{si2022grasp}. 

However, the tactile sim-to-real gap remains a significant obstacle~\cite{zhao2020sim}, limiting the direct application of tactile models learned in simulation to real-world environments. Bridging this gap has been the focus of various research efforts. For example, Narang et al. utilized 3D Finite Element models of BioTac sensors to simulate sensor deformations~\cite{narang2021interpreting}, enabling accurate regression of tactile sensor outputs but at the expense of computation cost. Similarly, Bi et al. achieved zero-shot sim-to-real transfer of a complex policy learned via RL by applying task-specific approximations for tractable simulation \cite{bi2021zero}. However, these approximations come at the cost of generalizability, requiring fine-tuning when transitioning to real-world tasks. Church et al. addressed this by incorporating real-to-sim pipelines to reduce the sim-to-real gap but noted that performance was hindered by the absence of key physical properties such as shear displacement in the simulation~\cite{church2022tactile}.

This work expands on the tactile sim-to-real method described in \cite{church2022tactile}, which uses a PyBullet simulator, depth image rendering, and pix2pix translation for zero-shot policy transfer. This approach has been applied successfully to a variety of tactile tasks—object pushing, edge following, surface following, and object rolling—on TacTip, DIGIT and DigiTac tactile sensors \cite{yang2023sim, lin2022tactile, church2022tactile, lin2023bitouch, lin2023attention}. However, none of these applications leveraged shear information. Our work preserves the computational efficiency of this methodology, while explicitly incorporating global shear information—enabling tasks that require a more detailed representation of the effect of lateral contact forces.

Further, these tactile sim-to-real approaches train models on simulated tactile images, and thus require a real-to-sim translation of incoming real tactile images at every inference step. Because the underlying simulator does not explicitly model shear, learned policies cannot exploit lateral force information. In contrast, our work adopts a genuine sim-to-real approach: we generate realistic tactile data that incorporates shear information. Policies can now be trained directly on these shear-enabled tactile images, eliminating the need for real-to-sim translation at runtime, and enabling models trained in simulation to sense and respond to shear in the real world.

\section{Methods}

\subsection{Tactile Robot System}


One Dobot MG400 4-axis desktop robotic arm, referred to as the follower robot, was equipped with a marker-based (331-pin) TacTip vision-based tactile sensor \cite{ward2018tactip}. This biomimetic optical tactile sensor has the benefit of giving tactile images that are sensitive to the distributions of both shear and normal forces during tactile interactions, enabling precise detection of surface deformations and fine object manipulation. A second MG400 arm, referred to as the leader robot, was not equipped with a tactile sensor and served to guide the movements of the object being tracked.

During training, only the follower robot was simulated for data collection. In contrast, our experimental setup uses both arms in a collaborative task, where the leader robot manipulates an object and the follower robot maintains continuous tactile contact by leveraging shear information. 

The experiments required access to the low-level servo controller of the robot arm, which is provided by the robot's manufacturer as an optional feature (as the robot is intended primarily for point-to-point position control). The gains of the rotatory joint of the end effector were tuned to be 2X stiffer to be less reactive to contact disturbances during the task. The servo control module provided by Dobot in their Python API allowed us generate smooth trajectories upon feeding successive poses to the arm, operating at around 10 cycles/second from the tactile image capture and processing.


\subsection{Tactile Simulation}



We use Tactile Gym 2.0 \cite{lin2022tactile}, a simulation environment designed specifically for tactile robotic manipulation tasks. Tactile Gym builds upon the PyBullet physics engine to quickly generate simulated tactile data. PyBullet renders depth images of interactions between the tactile sensor and objects, simulating the tactile feedback experienced by a sensor in contact with objects. These rendered depth images serve as our ``simulated tactile images'' using the method by Church et al. (2022)~\cite{church2022tactile}. 

As the simulated tactile sensor is approximated with rigid body physics, shear deformations of the tactile sensor are not modeled, and thus will never be present in simulated tactile images. However, we can infer shear displacements by extracting positional and rotational displacements from the simulator. When the sensor tip moves while in contact with an object, the shift in the contact surface's position and orientation relative to the sensor is used to generate a 6-dimensional ``shear pose'' (here 4 dimensional to match the DoFs of the robot arm), containing both positional and rotational shear components that complement the purely indentation depth-based tactile images. 

\subsection{Data Collection}

The collected datasets consist of tuples containing real tactile images, corresponding simulated tactile images, and a vector encoding the shear displacements derived from the physics simulator (visualization of data collection process shown in \autoref{fig:hero}). The robot performed a series of controlled interactions with the surfaces and edges of various objects, replicating a range of realistic contact scenarios.  We gathered 5000 tuples for training and 2000 for validation. Data collection took 130 minutes using our physical robot and 60 seconds in simulation. 


Although prior research has introduced shear-based movements primarily as a form of noise to increase the robustness of real-to-sim mappings \cite{lepora2021pose}, we incorporate shear as an essential facet of our data. By including explicit shear information, we enable a one-to-one sim-to-real image translation framework that accurately models how shear deformations affect the real tactile images. 

\subsection{Conditional U-Net GAN for Image-to-Image Translation}
\label{ssec:unetgans}

\begin{wrapfigure}{R}{0.55\textwidth}
    \vspace{-1.5em}  
    \centering
    \includegraphics[ trim=90 60 85 30, clip, width=.55\textwidth]
{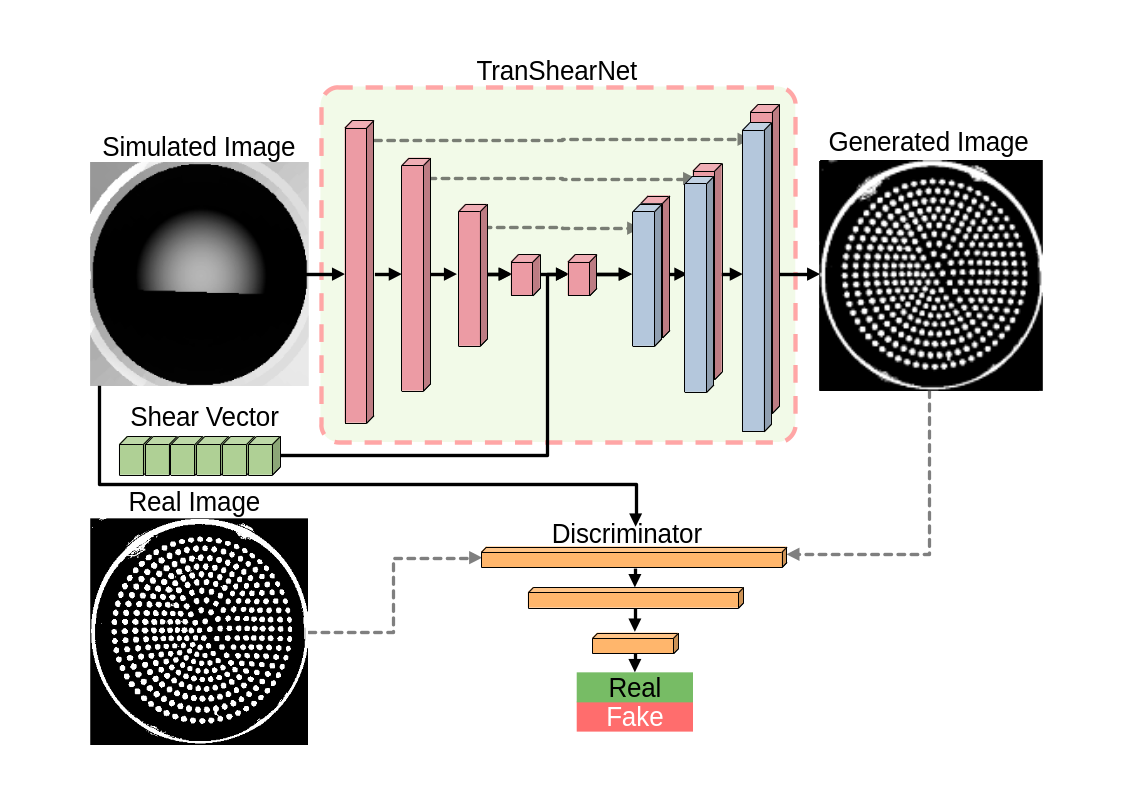}
    \caption{Our sim-to-real translation of tactile images uses a shPix2pix network: a modified pix2pix-trained GAN combined with a vector containing shear information in a fully connected layer.}
\label{fig:sim-to-real-arch}
    \vspace{-1.5em}  
\end{wrapfigure}



Previous work (e.g.~\cite{lin2022tactile,church2022tactile}) has used the pix2pix framework~\cite{isola2017image}, a widely adopted image-to-image translation model, to transfer real tactile images into their simulated counterparts. Pix2pix is a generative adversarial network (GAN) that consists of two main components: a U-net-based generator \cite{ronneberger2015u}, and a convolutional discriminator with batch normalization \cite{ioffe2015batch}. The U-net architecture is suited for tasks needing spatial information, as it uses skip connections to combine low-level feature maps from the downsampling layers with high-level feature maps from the upsampling layers \cite{ronneberger2015u}. 

Here we extend the vanilla pix2pix architecture to incorporate shear information explicitly. While previous methods~\cite{lin2022tactile,church2022tactile} used the standard U-net architecture to address the tactile sim-to-real gap, they did not account for shear information due to the limitations of rigid-body simulations. 
The U-net architecture cannot simulated tactile images into realistic ones due to the many-to-one relationship between simulated and real data.  
To address this, we add a fully connected layer with ReLU activation between the encoding (downsampling) and decoding (upsampling) layers of the U-net. The shear vector, which encodes both positional and rotational shear, is appended to the encoded representation before being passed through this fully connected layer (see \autoref{fig:sim-to-real-arch}). 

Our proposed framework enables the model to generate realistic tactile images that include shear deformations, even though simulated tactile images themselves lack shear information. By explicitly encoding shear into the representation, the model can generate tactile images that reflect real-world sensor behavior under contact and shear conditions. 

Our conditional U-net GANs were trained using paired datasets of simulated and real tactile images, along with shear vectors. The training process aimed to minimize the mean average pixel error (MAPE) for pixel-wise reconstruction between generated and real tactile images while simultaneously optimizing adversarial loss from the discriminator. Training was conducted for 100 epochs using a batch size of 16 and Adam optimizer \cite{kingma2014adam} with a learning rate of 0.0001 with early stopping. 

\subsection{Training Generalizable Contact Pose Estimators}

Previous work estimated tactile pose based on real tactile images \cite{lepora2021pose,lepora2020optimal}. Such research has used CNN architectures to predict contact pose from tactile images, which proved sufficiently effective to perform tactile servoing tasks such as robotic pushing, following, and surface following.

However, there are subtleties applying these methods to predicting shear information \cite{lloyd2021goal}. Slippage under shear can cause tactile aliasing~\cite{lloyd2021probabilistic} whereby similar sensor images become associated with very different poses and shears, which can lead to poorly performing models for predicting pose-and-shear. To address this issue, recent advancements have utilized a Gaussian-density neural network (GDNN) combined with a discriminative Bayesian filter \cite{lloyd2023pose}. The GDNN is particularly effective in modeling the continuous nature of pose and shear variables, enabling the network to output probability distributions over outputs, rather than discrete predictions, as shown in \autoref{fig:hero}. 

In this paper, we extend this pose-and-shear decoding approach to simulated tactile images generated using our trained shPix2pix networks: the conditional U-Net GANs described in \autoref{ssec:unetgans}. We trained Gaussian-density neural networks to decode both contact pose and shear information from our generated tactile images. By successfully training on simulated images, these models can provide accurate pose and shear estimates on real tactile images despite the absence of direct real-world data during training.

Our Gaussian-density neural networks were trained using the ShPix2pix generated datasets. The training data included U-net generated tactile images and corresponding pose and shear vectors. The networks were optimized using a negative log-likelihood loss over 50 training epochs with batch size of 64, using the Adam optimizer \cite{kingma2014adam} with a learning rate of 0.0001 and early stopping. 

\subsection{Task Formulation}

We evaluate our improved Sim-to-Real pipeline using two collaborative tasks \cite{lin2023bitouch, lin2018dual} that require both pose and shear-based tactile sensing. In both scenarios, two Dobot MG400 robotic arms are employed: one acts as the leader robot, while the other (the follower robot) is equipped with a biomimetic optical tactile sensor that uses a trained GDNN shear-estimation network.

\paragraph{Tactile Tracking Task:} In this task, the leader robot manipulates and rotates an object along a pre-programmed trajectory, and the follower robot actively tracks the object’s surface while preserving continuous contact as it moves and rotates in three-dimensional space. 

\paragraph{Collaborative Co-Lifting Task:} Both robots hold the object together, with the leader robot moving it along a specified path while the follower robot maintains a secure grip. As the load and object position changes, shear information becomes essential for preventing slips and dropping the object. 

In both tasks, reliable pose and shear estimation is essential. Pose information provides details on the object's relative position to the tactile sensor, while shear information is necessary to track sideways movements tangential to the sensor surface. Without shear data, the tactile robot would be unable to correct for those movements, and would quickly lose contact with the object.



\section{Results}

\subsection{Sim-to-Real Image translation}

\begin{wraptable}{R}{0.45\textwidth}
    \vspace{-5em}
    \centering
    \scriptsize
    \begin{tabular}{@{}l@{}c@{\hspace{1em}}c@{\hspace{1em}}c@{\hspace{1em}}c@{}}
    \toprule
    \multirow{2}{*}{\textbf{Method}} & \multicolumn{2}{c}{\textbf{Edge}} & \multicolumn{2}{c}{\textbf{Surface}} \\
    \cmidrule(lr){2-3}\cmidrule(lr){4-5}
    & \textbf{MAPE} $\downarrow$ & \textbf{SSIM} $\uparrow$ & \textbf{MAPE} $\downarrow$ & \textbf{SSIM} $\uparrow$ \\
    \midrule
    \textbf{pix2pix (baseline)} & 0.21 & 0.20 & 0.23 & 0.14 \\
    \textbf{shPix2pix (ours)}   & \textbf{0.07} & \textbf{0.63} & \textbf{0.11} & \textbf{0.68} \\
    \bottomrule
    \end{tabular}
    \caption{Sim-to-real image translation metrics. Bold values indicate the best results.}
    \label{tab:sim-to-real-results}
    \vspace{-1.5em}
\end{wraptable}

First, we evaluated our shPix2pix image translation network compared to a baseline vanilla pix2pix architecture. The vanilla pix2pix framework struggled to translate simulated tactile images into realistic ones due to the many-to-one relationship between simulated and real data. The baseline pix2pix architecture achieved a mean average pixel error (MAPE) of 0.22 and a structural similarity index measure (SSIM) of 0.17 when translating from simulated to real images.

Our shPix2pix image translation network successfully generated realistic tactile images that closely matched those obtained from the real tactile sensor. Explicitly encoding shear information into the model representation gives a significant reduction in MAPE to 0.091 and SSIM increase of 0.65.

\begin{wrapfigure}{R}{0.5\textwidth}
    \vspace{-1em}  
    \centering
    \includegraphics[width=0.5\textwidth]{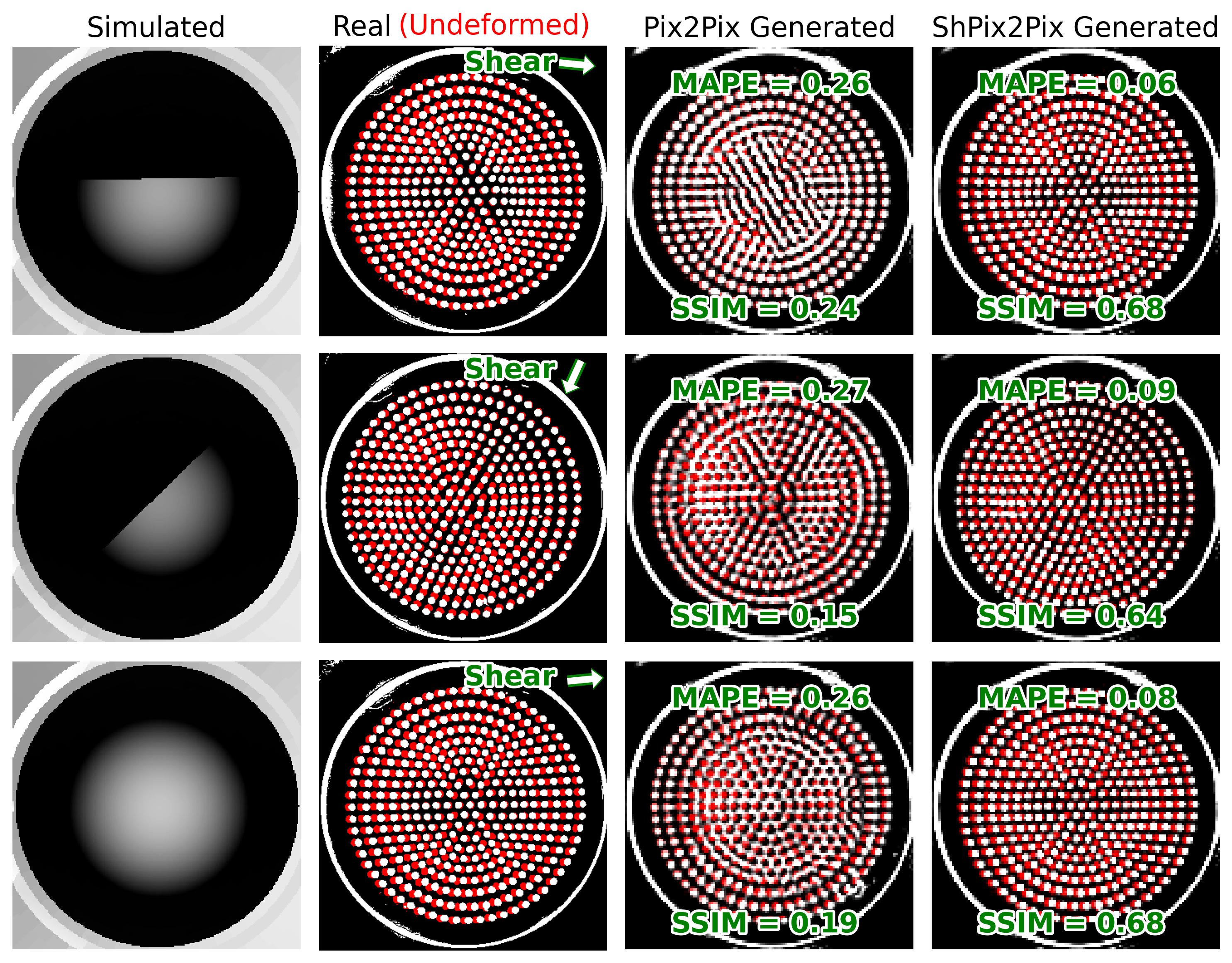}
    \vspace{-2em}  
    \caption{Comparison of sim-to-real tactile images. Undeformed tactile images are underlaid in red for reference against real/generated images.}
    \label{fig:sim-to-real-comparison}
    \vspace{-2em}  
\end{wrapfigure}

To further investigate this improvement, we examined two contact types within the dataset: \emph{Edge} contacts (where the sensor touches an edge of an object) and \emph{Surface} contacts (where the sensor touches a flat surface of an object). As shown in Table~\ref{tab:sim-to-real-results}, our shear-aware method consistently outperforms the baseline vanilla pix2pix network across both contact types.

\autoref{fig:sim-to-real-comparison} visualizes the results of this experiment, comparing the tactile images generated by the vanilla pix2pix model to those generated using our modified architecture. Our results demonstrate that the integration of the shear vector into the conditional U-Net GAN enables more effective sim-to-real tactile image translation over varying contact conditions such as edges and surfaces.


\subsection{Sim-to-Real PoseNet}

\begin{figure}[b!]
    \centering
        \begin{tabular}{@{}cc@{}}
            \textbf{vanilla pix2pix (baseline)} & \textbf{shPix2pix (ours)}\\
            \includegraphics[trim=0 0 0 0, clip, width=0.49\columnwidth]{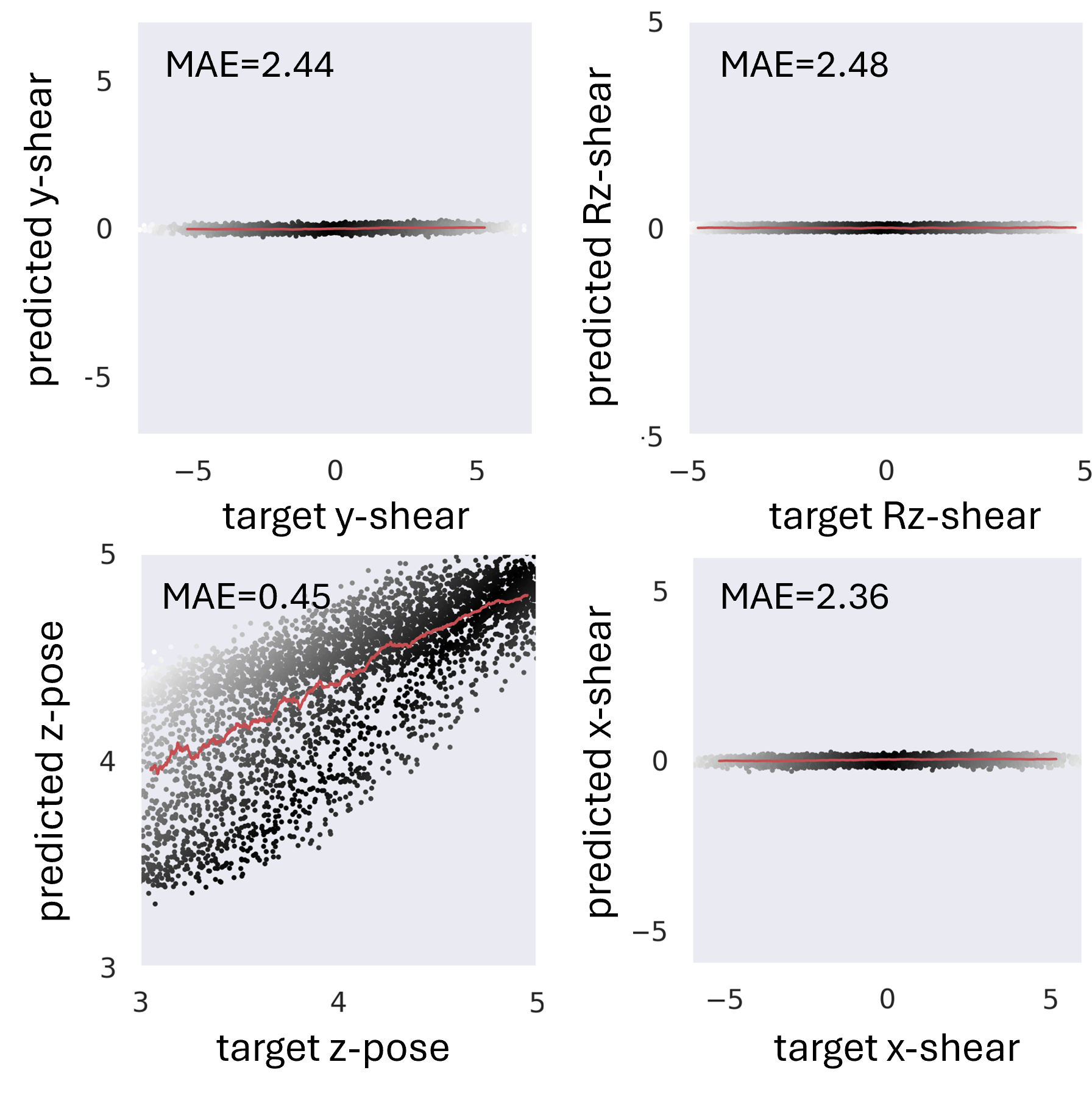}&
            \includegraphics[trim=0 0 0 0, clip, width=0.49\columnwidth]{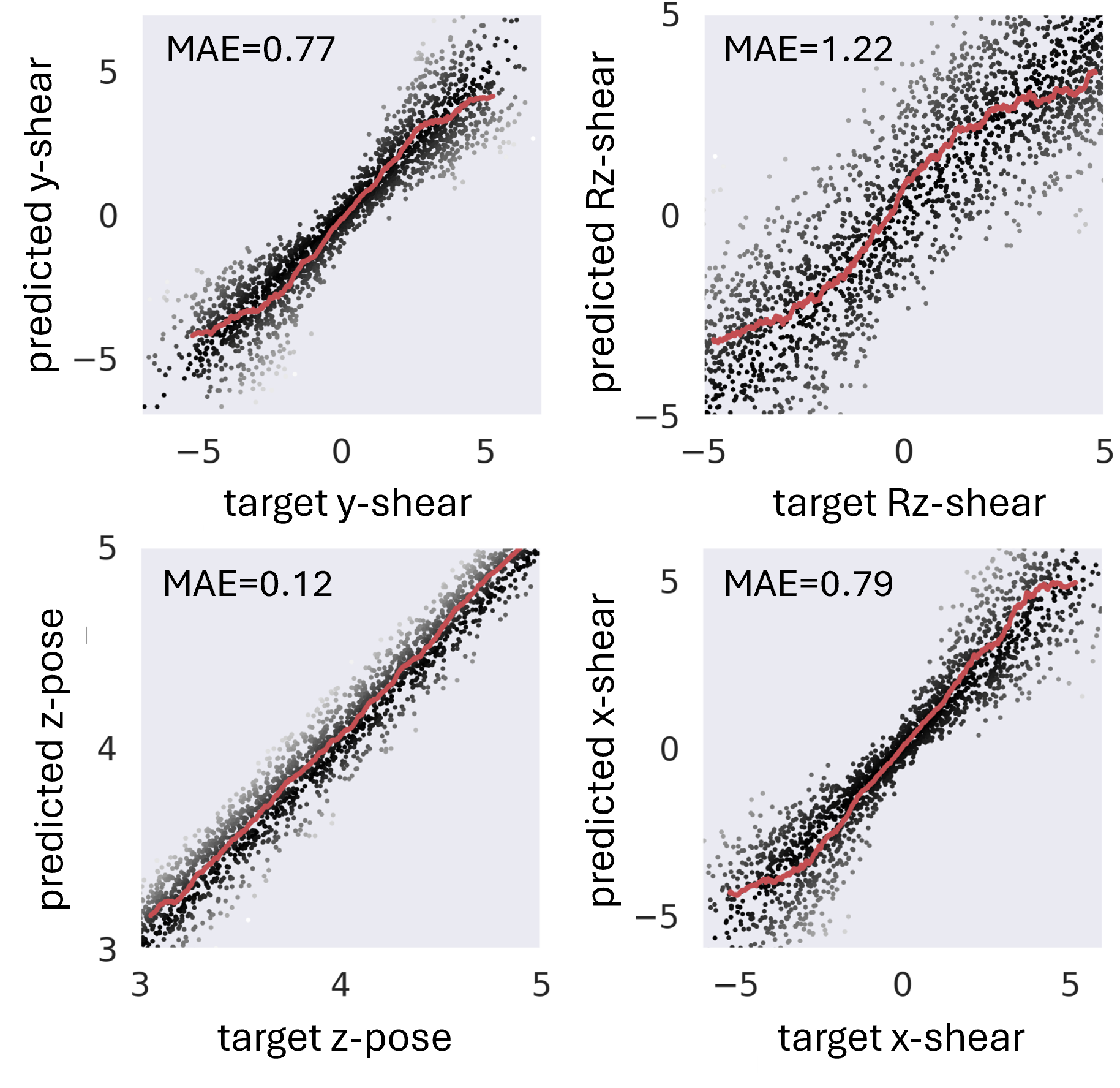}\\
        \end{tabular}
        \caption{Shear- and pose-prediction errors for Gaussian-density neural networks trained using the baseline pix2pix and our proposed shPix2pix sim-to-real data generation methods.}
    \label{fig:posenet-decoding-error}
\end{figure}

To evaluate our Sim-to-Real pipeline, we trained Gaussian-density neural networks using data generated from the image translation methods. In the baseline approach, we generated training data by translating simulated tactile images to the real domain with a standard pix2pix model. In the second approach, we utilized our shPix2pix architecture, which incorporates the shear vector to generate tactile images containing realistic shear deformations. Both networks were tested exclusively on real tactile images, to assess their generalization to real-world conditions.

Our results indicate that while models trained on pix2pix-generated images can decode pose with above-random success, they are unable to infer shear information (\autoref{fig:posenet-decoding-error}, left). In contrast, the models trained on shPix2pix images are able to accurately predict both pose and shear variables on real data never encountered during training (\autoref{fig:posenet-decoding-error}, right). Note that the prediction errors for shear in the shPix2pix-trained model are comparable to those reported for networks trained entirely on real tactile data \cite{lloyd2023pose} (i.e., without sim-to-real), highlighting the robustness of our approach.




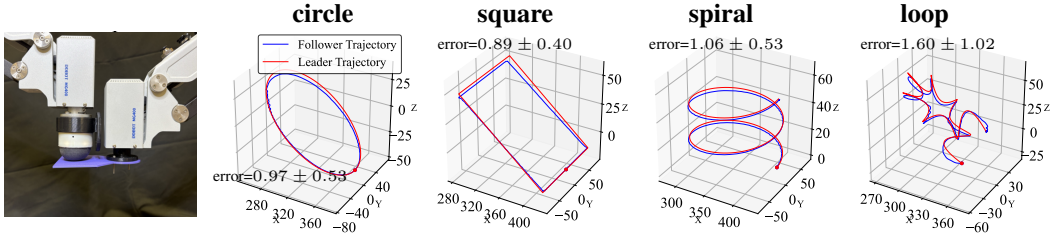
\begin{figure*}[h!]
\vspace{-2em}
\begin{picture}(100,100)
\put(0,0){\includegraphics[trim=0 0 0 65, clip, width=\columnwidth]{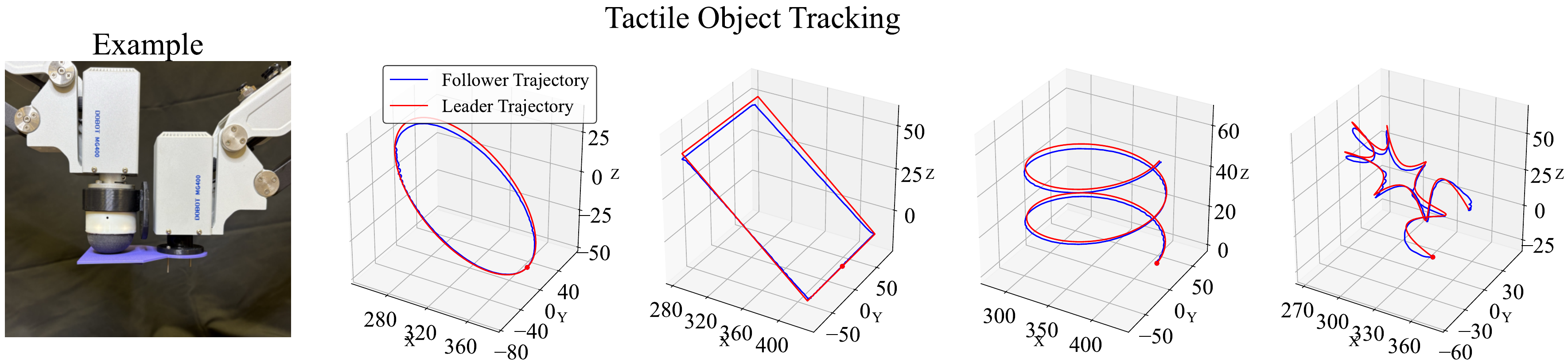}}
\put(110,80){\bf circle}\put(80,20){\tiny error=$0.97\pm0.53$}
\put(180,80){\bf square}\put(165,70){\tiny error=$0.89\pm0.40$}
\put(260,80){\bf spiral}\put(245,70){\tiny error=$1.06\pm0.53$}
\put(340,80){\bf loop}\put(325,70){\tiny error=$1.60\pm1.02$}
\end{picture}
\caption{Tactile object tracking task. Left: experimental setup. The planar purple surface is mounted as the leader's end effector.  Right: motions of the leader (red) and tactile follower (blue) robots under four distinct trajectories varying in object shear and pose. We refer also to the video results included in supplemental material.}
\label{fig:tracking-trajectories}
\vspace{-.5em}
\end{figure*}

\subsection{Tactile Object Tracking}
\label{ssec:tac_track}



Next, we tested the tactile robot's ability to follow a leader robot across various complex trajectories: circles, tilted squares, upward spirals, and a complex loop. The follower MG400 successfully maintained continuous contact with the moving object over all tested trajectories. On average, the distance between the tactile sensor’s position and the target position on the object’s surface was 1-2\,mm, as marked by the error on the trajectory plots (with standard deviation over a single trajectory, indicating this error was consistent). Repeated runs of the same experiment gave the same results, with the `looping' trajectory most difficult because of the complex shape. Visually, the tracked trajectories of the tactile follower closely matched the trajectory of the leader robot~(\autoref{fig:tracking-trajectories} and supplemental video), confirming the precision of the sim-to-real tactile servo controller.

Note that the GDNN models used to estimate pose and shear information for this task were trained entirely on simulated tactile images generated by our shPix2pix conditional U-net GAN pipeline. The servo control system achieved robust performance in real-world conditions despite having never encountered real tactile images captured by real tactile sensors during training. 


\subsection{Collaborative Co-Lifting Task}
\label{sec:co-lift}

Finally, we tested the ability of the follower and leader robots to jointly lift and hold various objects while the leader moved it along two trajectories (a curved circular wave and a pointed star). The choice of objects were chosen to show generalization performance: (1) a rigid square prism similar to the sim-to-real training data; (2) a rigid egg with curved surface; (3) a soft `squidgy' brain; (4) a curved ducky. The follower robot was now equipped with a horizontally mounted tactile sensor, to also show generalization performance under a change of gravity direction. 

\begin{figure*}[]
\vspace{18em}
\begin{picture}(100,100)
\put(0,0){\includegraphics[trim=0 0 0 80, clip, width=\columnwidth]{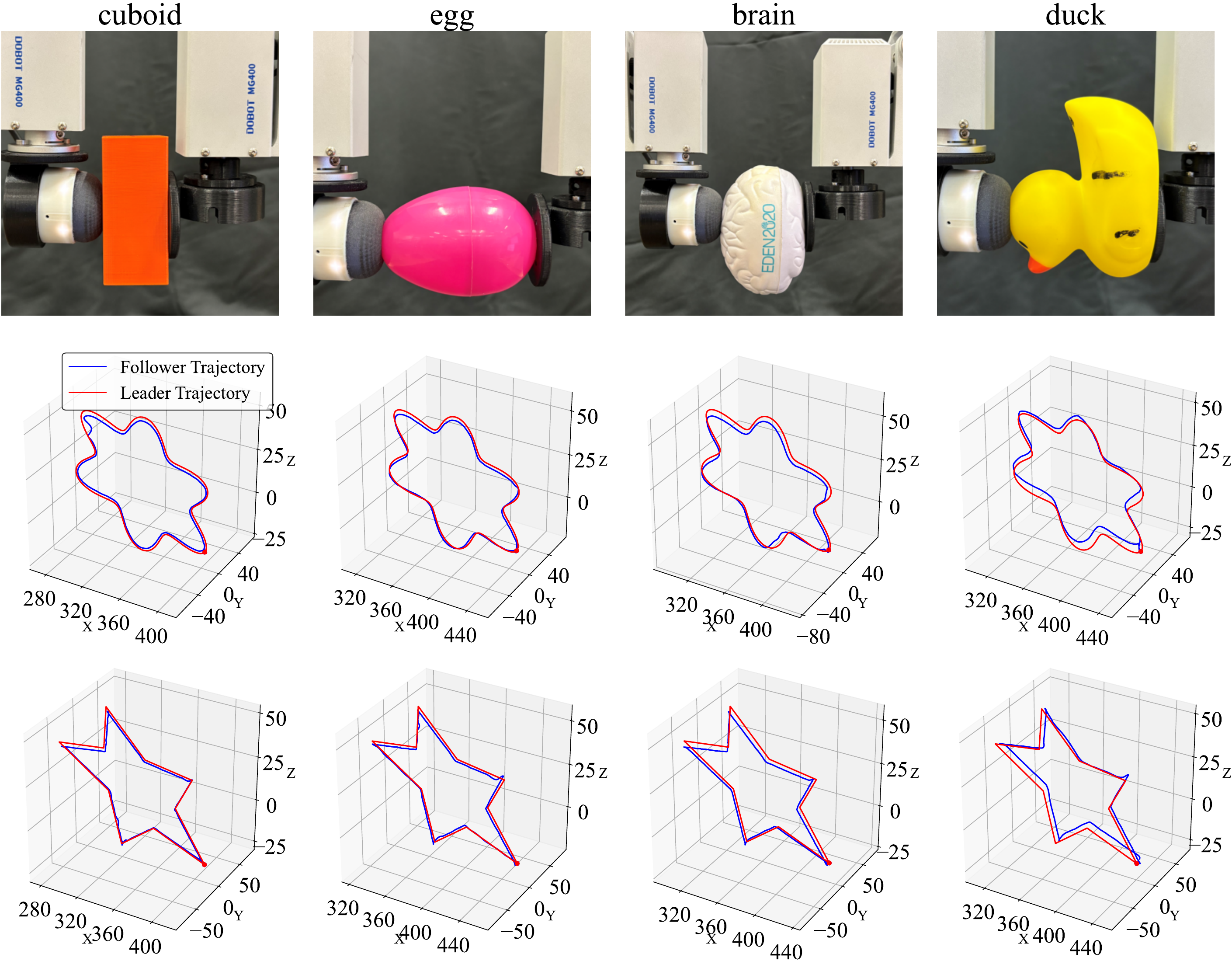}}
\put(20,290){\bf square prism}\put(10,25){\tiny error=$1.01\pm0.53$}\put(10,125){\tiny error=$1.42\pm0.82$}
\put(130,290){\bf rigid egg}\put(120,85){\tiny error=$1.48\pm0.69$}\put(120,185){\tiny error=$1.45\pm0.70$}
\put(230,290){\bf soft brain}\put(220,85){\tiny error=$1.96\pm0.97$}\put(220,185){\tiny error=$2.44\pm1.55$}
\put(330,290){\bf ducky}\put(320,85){\tiny error=$2.06\pm0.49$}\put(320,185){\tiny error=$1.95\pm1.34$}
\end{picture}
\caption{Collaborative object lifting task. Top row: experimental setup with the four distinct curved and soft objects. Middle and bottom rows: motions of the leader (red) and tactile follower (blue) robots under distinct trajectories varying in object shear and pose. That these trajectories are a close match led to a secure grasp. We refer also to the video results included in the supplemental material.}
\label{fig:co-lift-trajectories}
\vspace{-1em}
\end{figure*}
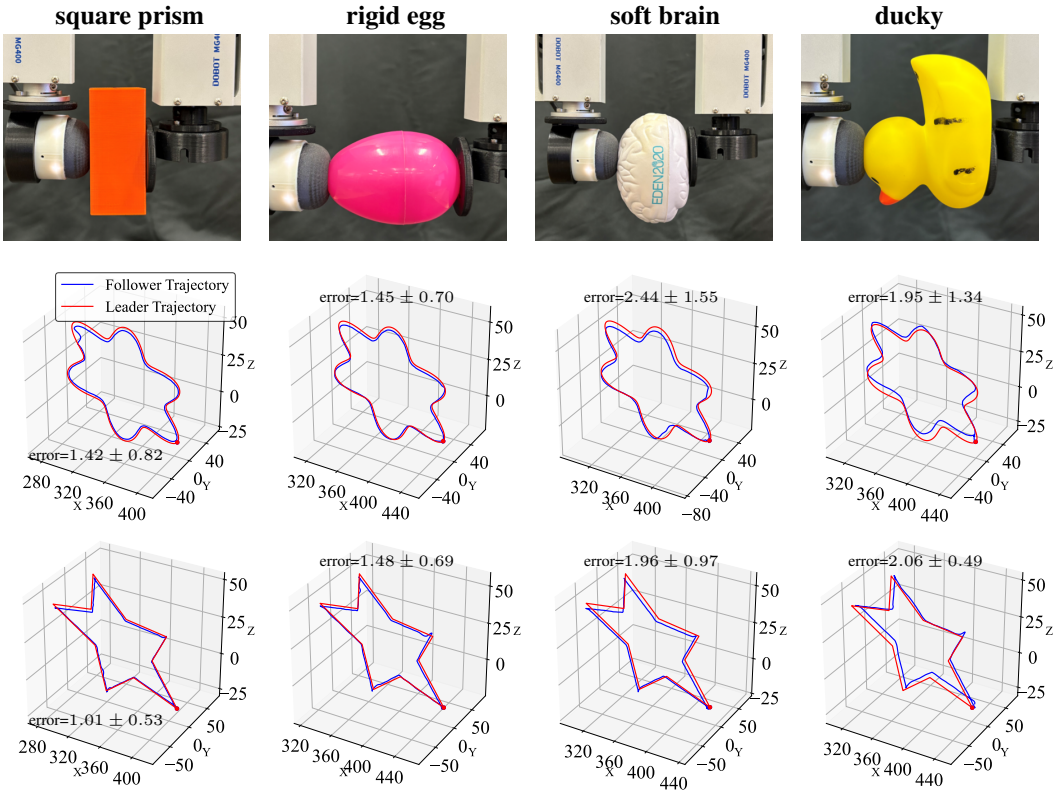

In all tested scenarios, the tactile follower robot maintained continuous contact that was sufficiently precise in following the leader robot to securely hold the object while it was being moved along the leader' trajectory. Again, the trajectory errors were in the range 1-2\,mm, consistent with a close visual match between the leader and follower trajectories (\autoref{fig:co-lift-trajectories} and supplemental video). The most challenging object was the soft brain, because the soft contact led to deformation of the tactile images and less reactive control, although good performance was still maintained. The ducky was also challenging, which we attribute to its larger weight, as we did not tune the system to compensate for the differences in vertical shear caused by the differing object weights.  


Again, we note that the follower robot’s pose and shear estimation network was trained exclusively using shPix2pix-generated data, without any training examples featuring horizontally oriented tactile sensors, co-lifting tasks, curved surfaces or soft objects. Yet, the co-lift maneuvers proceeded smoothly and securely, with the follower robot dynamically adjusting its position and grip to match the leader’s movements communicated solely through the changes in object pose and shear. This outcome highlights the generalization capability of our SimShear approach for shear-aware tactile manipulation on a demanding task requiring precision for success in situations distinct from training.


\section{Conclusion} 
\label{sec:conclusion}

We introduced SimShear, a novel Sim-to-Real pipeline for tactile sensing that integrates shear information to enhance the accuracy and versatility of robotic manipulation tasks trained in simulation. By employing a conditional U-Net GAN, the shPix2pix, our method overcomes the limitations of current Real-to-Sim pipelines that cannot model the effects of shear force. Our method removes two primary drawbacks of current tactile real-to-sim pipelines—namely, the lack of shear in simulation and the need to translate real images into the simulated domain at every inference step. Instead, we generate realistic, shear-enabled tactile data for policy training, allowing robots trained in simulation to sense and respond to lateral displacements directly in real-world scenarios.

These findings suggest that incorporating shear information into the Sim-to-Real pipeline not only delivers more precise manipulation but also facilitates a broader range of tactile control tasks—extending the potential of Sim-to-real tactile sensing into more challenging applications. By removing the need for real-to-sim image translation at every step, our method offers a simpler and more efficient framework for deploying tactile policies trained entirely in simulation. Moving forward, this approach could be adapted to more complex tasks and diverse sensor configurations, enabling the advancement of tactile robotic dexterity through improved sim-to-real transfer.  

\section{Limitations} 
\label{sec:conclusion}

We demonstrated the effectiveness of SimShear by showing how a Gaussian-density neural network trained on shPix2pix-generated data can accurately estimate both pose and shear. This capability was validated in two tasks where precise reactions to changes in shear are essential. First, a tactile tracking task, where a follower manipulator maintains continuous horizontal contact with a moving object guided by a leader robot. Second, a collaborative co-lifting task, where the leader and follower robots jointly hold and move an object along complex trajectories. 

The most significant limitation was that our choice of desktop robot (the Dobot MG400) constrained the experiments to the 4 degrees of freedom of the robot, where the end effector can only move position and rotate around the vertical. Therefore, we were limited to the two tasks considered (tracking and co-lifting), which we transitioned between by physically remounting the sensor from being vertically- to horizontally-oriented. Also, the models were limited to 4 pose and shear variables (\autoref{fig:posenet-decoding-error}). We made this choice of robotic system because it was considered sufficient to demonstrate our SimShear approach, and we believe it is important to use an affordable system that is accessible to others. That said, future work will consider the extension of the SimShear approach to higher degree-of-freedom manipulators such as 7-DoF robot arms for a greater range of tasks (e.g. those of reference~\cite{lloyd2023pose}, but using sim-to-real). The challenge to solve in that situation is to generalize the models to 6 rather than 4 pose-and-shear variables and ensure accuracy and generalization of the models across a broader range of tasks. 

Another limitation was that we considered only fairly light objects in the co-lifting task. This choice was due to complexities in needing to compensate for the vertical shear caused by object weight, which for a range of heavier objects would require individual tuning per object. This can be done easily, e.g. by tuning the tactile servo controller to not drift downwards when holding each object stationary, but we felt this was a diversion from the main subject of the paper on shear-based sim-to-real control. If more dynamic tasks are considered in the future, where the robots move rapidly and the contact strength changes, this inertial correction could become more complicated and require a motion-based model to compensate.

Even though our approach using a shPix2pix network showed high generalization, being trained on data from an experimental configuration that differed significantly from the actual test scenarios, the generalization capabilities could be extended further. Currently, we considered  differences in tactile sensor orientation and with curved and/or soft contact shapes, compared to flat rigid objects in training. This could be extended to objects with sharp or pointed contact surfaces, or very soft or slippery objects, which the current models would not be able to accurately predict the pose and shear. We expect this extension could be achieved by extending the sim and real training data set for the shPixpix network, which would also enable a much broader range of tasks to become possible with SimShear. In the future, we expect such extensions could encompass many manipulation actions achievable through coordinating multiple tactile robots via contact with the manipulated object. We will also explore spatial RL data agumentation techniques (e.g., \cite{laskin2020reinforcement, lin2020iter}) in combination with SimShear-generated tactile images to further enhance learning generalizability.


\acknowledgments{NL and YL were supported by the Advanced Research + Invention Agency (`Democratising Hardware and Control For Robot Dexterity'). We thank the reviewers and area chair for the constructive feedback given on this paper.}


\bibliography{example}  

\end{document}